%% file: neurips_2025.tex
\documentclass{article}

\PassOptionsToPackage{numbers, sort&compress}{natbib}
 
\usepackage[creativeai,final]{neurips_2025}

\usepackage[utf8]{inputenc} 
\usepackage[T1]{fontenc}    
\usepackage{hyperref}       
\usepackage{url}            
\usepackage{booktabs}       
\usepackage{amsfonts}       
\usepackage{nicefrac}       
\usepackage{microtype}      
\usepackage{xcolor}         
\usepackage{graphicx}
\usepackage{amsmath}

\usepackage{multirow}
\usepackage{float}
\usepackage{siunitx}

\title{Exploring Human-AI Conceptual Alignment through the Prism of Chess}

\author{%
  Semyon Lomasov\textsuperscript{*}$^{1}$,\And
  Judah Goldfeder\textsuperscript{*}$^{2}$,\And
  Mehmet Hamza Erol\textsuperscript{*}$^{1}$,\And
  Matthew So\textsuperscript{*}$^{2}$,\And
  Yao Yan$^{3}$,\And
  Addison Howard$^{3}$,\And
  Nathan Kutz$^{4}$,\And
  Ravid Shwartz Ziv$^{5}$
  \AND
  \\
  $^{1}$Stanford University \quad
  $^{2}$Columbia University \quad
  $^{3}$Kaggle \quad
  $^{4}$University of Washington \quad
  $^{5}$NYU
}

\begin{document}
\maketitle

\begingroup
\renewcommand\thefootnote{\fnsymbol{footnote}}
\footnotetext[1]{Equal contribution.}
\endgroup

\begin{abstract}

Do AI systems truly understand human concepts or merely mimic surface patterns? We investigate this through chess, where human creativity meets precise strategic concepts. Analyzing a 270M-parameter transformer that achieves grandmaster-level play, we uncover a striking paradox: while early layers encode human concepts like center control and knight outposts with up to 85\% accuracy, deeper layers, despite driving superior performance, drift toward alien representations, dropping to 50-65\% accuracy. To test conceptual robustness beyond memorization, we introduce the first Chess960 dataset: 240 expert-annotated positions across 6 strategic concepts. When opening theory is eliminated through randomized starting positions, concept recognition drops 10-20\% across all methods, revealing the model's reliance on memorized patterns rather than abstract understanding. Our layer-wise analysis exposes a fundamental tension in current architectures: the representations that win games diverge from those that align with human thinking. These findings suggest that as AI systems optimize for performance, they develop increasingly alien intelligence, a critical challenge for creative AI applications requiring genuine human-AI collaboration. Dataset and code are available at: \url{https://github.com/slomasov/ChessConceptsLLM}.

\end{abstract}

\section{Introduction}

When Garry Kasparov faced Deep Blue in 1997, he described moments where the machine's moves seemed to exhibit genuine creativity. This raises a fundamental question for creative AI: Do these systems truly understand human concepts, or are they pattern-matching engines that coincidentally produce creative outputs?



Chess provides an ideal laboratory for this investigation, offering precise definitions for human concepts like \textit{center control} and \textit{king safety} while remaining a creative, strategic game. Recent transformer-based engines \cite{ruoss2024grandmaster} achieve grandmaster-level play without explicit search, raising a tantalizing possibility: perhaps these models learn to think in human-like concepts. However, their performance collapses on Chess960 (Fischer Random Chess), where pieces start in randomized positions. Although the same strategic concepts are present, this brittleness suggests these systems might be memorizing patterns rather than understanding principles, hinting at a deeper misalignment between human and machine cognition.





\textbf{Our Contributions:} We present the first systematic investigation of conceptual alignment between humans and neural chess engines through three key contributions:

\begin{enumerate}
    \item \textbf{A Novel Chess960 Dataset:} We introduce the first expert-curated dataset of 240 Chess960 positions, manually annotated with 6 fundamental chess concepts. This dataset enables testing whether AI systems grasp abstract concepts or merely memorize standard patterns, which is a crucial distinction for creative AI systems that must generalize beyond their training.
    
    \item \textbf{Multi-Method Concept Detection:} We develop three complementary approaches for probing concept representations in transformer layers, revealing how different architectures encode human knowledge. Our methods range from sparse concept vectors to neural network probes, providing robust evidence of conceptual (mis)alignment.
    
    \item \textbf{Layer-wise Alignment Analysis:} We uncover a surprising inversion: early transformer layers show strong alignment with human concepts (up to 85\% accuracy), while deeper layers, despite superior playing strength, drift toward alien representations. This suggests a fundamental tension between performance optimization and human-interpretable reasoning.
\end{enumerate}

Our findings reveal that while transformers capture surface-level chess concepts, this alignment is fragile and diminishes as models optimize for performance. When tested on Chess960, where memorized opening knowledge becomes useless, concept recognition accuracy drops significantly, exposing the shallow nature of learned representations.

These results have profound implications for creative AI: systems appearing to share our concepts may operate on fundamentally different principles, highlighting the need for architectures that maintain human alignment throughout their processing hierarchy.

\section{Related Work}

Our investigation bridges three research areas: neural approaches to chess, interpretability of creative AI systems, and human-AI conceptual alignment. We position our work at their intersection, addressing how AI systems develop representations that may diverge from human creative thinking.

\subsection{Neural Chess Engines and Interpretability}

 


The renaissance of neural approaches to chess began with AlphaZero \cite{silver2017masteringchessshogiselfplay}, which learned superhuman play through self-play without human knowledge. Recent work has pushed further: transformer-based engines now achieve grandmaster-level play without explicit search. Ruoss et al. \cite{ruoss2024grandmaster} developed the 270M-parameter model we analyze, which frames chess as a sequence prediction problem, achieving 2750 Elo on standard chess but dropping closer to amateur level on Chess960.

Contemporary approaches continue to evolve. Monroe and Chalmers \cite{monroe2024masteringchesstransformermodel} demonstrate the feasability of transformer models for chess, while Schultz et al. \cite{schultz2025masteringboardgamesexternal} explore combining external and internal planning with language models. Jenner et al. \cite{jenner2024evidence} found evidence of learned look-ahead in chess networks, suggesting some form of planning emerges. Most relevant to our work, Schut et al. \cite{schultz2025masteringboardgamesexternal} analyzed concept discovery in AlphaZero, examining how the system develops human-interpretable concepts. However, their analysis focused solely on standard chess positions. Our work extends this by applying the analysis to a searchless transformer model, and by introducing Chess960 as a critical test of conceptual robustness.



\subsection{Layer-wise Analysis and Interpretability}
Understanding how neural networks build representations across layers has become crucial for interpretability. Skean et al. \cite{skean2025layerlayeruncoveringhidden} provide a comprehensive layer-by-layer analysis revealing how hidden representations in language models evolve, showing that different layers capture fundamentally different types of information. This layer-wise specialization appears to be a general principle across domains. In the context of human-AI alignment, Shani et al. \cite{shani2025tokensthoughtsllmshumans} demonstrate that humans and language models compress information differently, with profound implications for how these systems understand concepts. They show that while models may achieve similar outputs, their internal representations follow alien optimization paths. Our chess analysis reveals a parallel phenomenon: early layers maintain human-aligned representations while later layers diverge toward performance-optimized but conceptually opaque strategies.

\subsection{The Creative Gap in AI Systems}
Our work reveals a fundamental tension in current AI architectures: the representations that achieve best performance diverge from those that align with human thinking. This has critical implications for creative AI applications where human-AI collaboration requires shared conceptual frameworks. By showing that conceptual alignment deteriorates as networks optimize for task performance, we highlight a key challenge for building AI systems that can serve as genuine creative partners rather than inscrutable optimization machines.


\section{Methods}

We develop a systematic framework to probe whether transformer-based chess models encode human strategic concepts. Our approach combines a novel Chess960 dataset with three complementary probing techniques, enabling robust measurement of conceptual alignment across model layers.

\subsection{Dataset Construction}
\subsubsection{Classical Chess Dataset}

We leverage the Strategic Test Suite (STS) \cite{corbit2014strategic}—1,500 expert-curated chess positions labeled with strategic concepts. We focus on six core concepts that translate across game phases: \textbf{Open Files and Diagonals}, \textbf{Knight Outposts}, \textbf{Advancement of f/g/h pawns} (kingside), \textbf{Advancement of a/b/c pawns} (queenside), \textbf{Center Control}, and \textbf{Pawn Play in the Center}. We exclude overlapping concepts and endgame-specific patterns where starting positions become irrelevant. For an example of what concepts look like, see Appendix \ref{app:conc}.

\subsubsection{Novel Chess960 Dataset}

To test conceptual understanding beyond memorization, we created the first annotated Chess960 dataset. Chess960 randomizes piece placement on the back rank while preserving chess rules, eliminating memorized opening theory while maintaining strategic principles. Our dataset contains 240 positions (40 per concept) curated from high-level Chess960 games and annotated by experts (>2200 Elo). Each position was selected to clearly exemplify its target concept while matching the complexity distribution of STS positions. This dataset enables us to answer: when opening memorization is impossible, do models still recognize fundamental strategic patterns?

\subsection{Model Architecture and Activation Extraction}

\begin{figure}[ht]
  \centering
  \includegraphics[width=\linewidth]{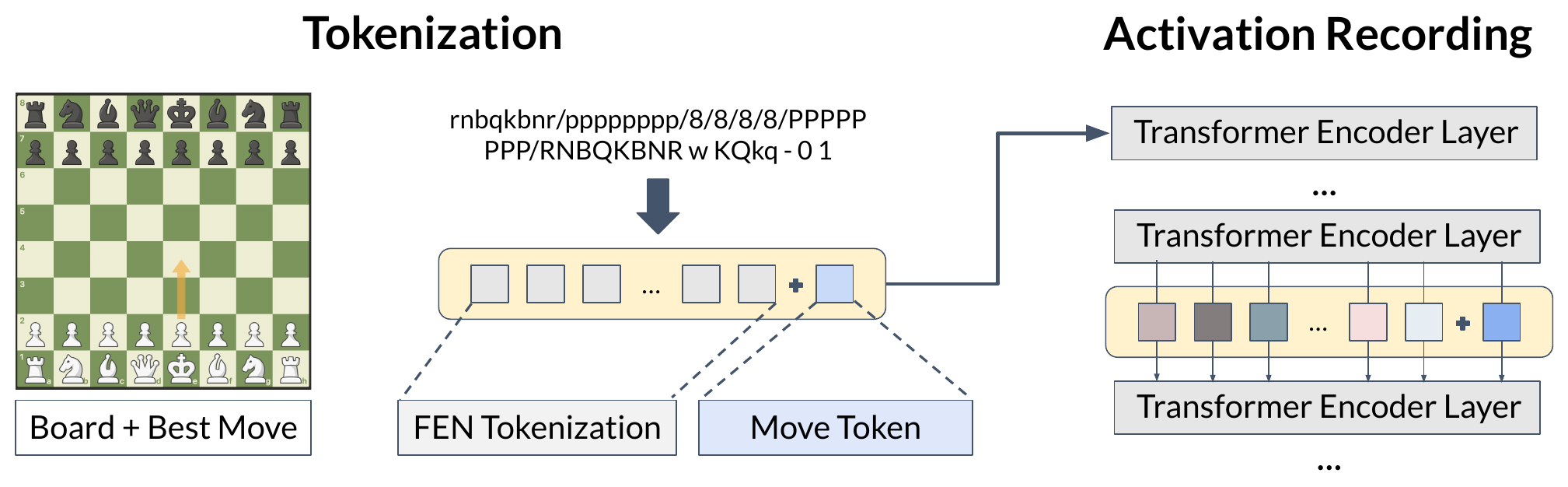}
\caption{\textbf{From board to move: tracking where human concepts disappear in the processing pipeline.} The model encodes positions as FEN strings, appends move tokens, and processes them through transformer layers. Recording activations at each layer reveals where strategic understanding shifts from human-recognizable patterns to alien representations.}

  \label{fig:model_pipeline}
\end{figure}

We analyze the 270M-parameter transformer from Ruoss et al. \cite{ruoss2024grandmaster}, which achieves grandmaster-level play on classical chess but struggles on Chess960. The model encodes positions as FEN strings, processes them through 18 transformer layers with 1024-dimensional embeddings, and outputs move probabilities (Figure~\ref{fig:model_pipeline}). For our analysis, we extract activations at each layer for the model's chosen move, yielding representations $z \in \mathbb{R}^{L \times T \times D}$ where $L=18$ layers, $T=79$ tokens, and $D=1024$ dimensions. These activations reveal how the model's internal representations evolve from raw position encoding to final move selection.

\subsection{Probing Methodologies}
We employ three complementary approaches to detect concept representations, each offering different insights into how the model encodes human knowledge.

At the core of our approach is a simple principle: if the model truly understands a concept, there should exist directions in its activation space that distinguish positions containing that concept from those without it. Formally, for a given layer's activation representation $r \in \mathbb{R}^D$, we seek a vector $v$ that maximizes:
$v^T r_{concept} > v^T r_{no\text{-}concept}$.

This intuition drives all three methods. \textbf{Sparse Concept Vectors} find the minimal set of neurons (using L1 regularization) that distinguish concepts, revealing which specific dimensions encode strategic knowledge. \textbf{Logistic Regression} learns linear decision boundaries with probabilistic outputs, providing interpretable separation between positions with and without each concept. \textbf{Sequence-Aware Neural Probes} extend this to process all token activations jointly in a layer through a lightweight network, capturing distributed representations across the full position encoding. Full mathematical formulations for each method are provided in Appendix \ref{app:math}.

Each method offers complementary insights: sparsity reveals which neurons matter most, logistic regression provides interpretable boundaries, and neural probes capture complex patterns. Crucially, all three methods show consistent results: strong concept alignment in early layers that deteriorates in deeper layers, suggesting a fundamental tension between human-interpretable representations and performance optimization.

\subsection{Experimental Design}

We evaluate four experimental scenarios to comprehensively assess conceptual alignment. \textbf{Scenario I} establishes baseline performance by training and testing on classical chess positions. \textbf{Scenario II} tests generalization by training on classical chess but testing on Chess960, measuring robustness to distributional shift. \textbf{Scenario III} explores adaptation by training and testing on combined datasets. \textbf{Scenario IV} isolates Chess960 performance to measure concept learning without standard patterns. We analyze layers 2, 5, 10, and 15 to capture the full spectrum from early feature extraction to final move selection, performing 5-fold cross-validation for robust results.

\section{Results}
Our experiments reveal a striking paradox: as the transformer model becomes more capable at chess, it becomes less human-like in its thinking. This fundamental tension between performance and interpretability emerges consistently across all our analyses.

\subsection{The Fragility of Conceptual Understanding}

Table~\ref{tab:scenario-pivot} presents our central finding: human chess concepts are surprisingly fragile to disruption. When tested on standard chess positions (Scenario I), our probes achieve strong accuracy, with the Concept Vector method reaching 85.68\% for "Pawn Play in the Center." However, this apparent understanding deteriorates significantly when confronted with Chess960's creative disruption.

The most telling comparison lies between Scenarios I and II. When we train on standard chess but test on Chess960, accuracy drops consistently across all methods. Knight outpost recognition plummets from 83.60\% to 72.36\% with Concept Vectors, while logistic regression drops even more dramatically from 79.60\% to 67.92\%. This reveals that the model's "understanding" relies on memorized patterns rather than genuine conceptual knowledge.

Interestingly, the three probing methods show different robustness to disruption. Concept Vectors maintain 70-76\% accuracy even in pure Chess960 scenarios, while the Sequence-Aware Neural Probe often falls below 60\%, suggesting sparse representations transfer better than distributed patterns. Even "Center Control", geometrically identical across variants, shows degradation, indicating the model conflates abstract principles with specific configurations.

Most counterintuitively, training exclusively on Chess960 (Scenario IV) yields lower accuracy than zero-shot transfer from standard chess (Scenario II) for several concepts. Combined training (Scenario III) sometimes achieves the highest accuracy: "Pawn Play in the Center" reaches 86.92\%, but only when standard positions remain in the mix. These patterns suggest Chess960 positions are inherently harder for the model to parse conceptually, possibly because the absence of opening principles forces more complex evaluation from the start.








\input{best_layer_results}
\subsection{Layer-wise Divergence: The Alien Mind Emerges}
\label{sec:layer-wise-divergence}

\begin{figure}[ht]
  \centering
  \includegraphics[width=\linewidth]{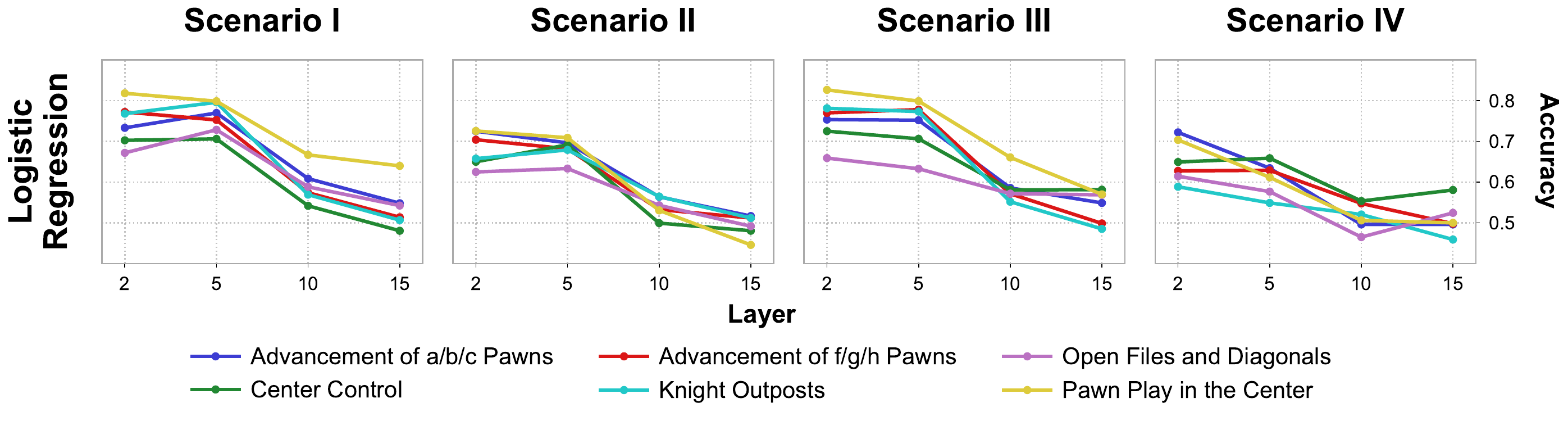}
  \caption{\textbf{Human concepts fade as the network goes deeper: early layers think like humans, late layers think like aliens.} Layer-wise accuracy for detecting six chess concepts using Logistic Regression probing. For a comparison of all 3 probing methods, see Appendix \ref{app:res}. Early layers (2-5) achieve 70-85\% accuracy, dropping to 50-65\% by layer 15 across all methods. Standard chess shows smooth degradation while Chess960 becomes erratic, revealing unstable representations without memorized patterns. This universal decline exposes the trade-off between human interpretability and performance.} 
  \label{fig:concept_trends}
\end{figure}

Figure~\ref{fig:concept_trends} unveils perhaps our most intriguing discovery: the model's journey from human-like to alien thinking happens gradually but consistently across its layers. In early layers (2-5), we observe robust concept detection across most scenarios—the Concept Vector method maintains accuracies around 80-85\% for well-defined concepts. These layers, close to the raw board encoding, maintain representations that align with how humans parse chess positions.

But as we probe deeper layers, we witness a universal decay in human concept alignment. By layer 15, accuracies typically drop to 50-65\% across most concepts, a decline of 15-30 percentage points from their early-layer peaks. This pattern is remarkably consistent across all three probing methods, though the Sequence-Aware Neural Probe shows more volatility and generally lower performance (often 10-15\% below the other methods), suggesting that full-sequence information becomes increasingly alien in its organization.

The degradation pattern varies revealing by scenario. Classical chess (Scenario I) shows smooth decline from layers 2 to 15, while Chess960 after classical training (Scenario II) drops sharply early then stabilizes at lower accuracy—the model struggles to apply learned concepts from the first layers. Combined training (Scenario III) mirrors classical chess with better retention, but pure Chess960 (Scenario IV) produces erratic patterns with high variance, suggesting that without memorized openings, the model develops unstable representations.

Notably, some concepts show dramatic late-layer collapse. For example, in Scenario II, "Pawn Play in the Center" using Concept Vectors drops from ~75\% at layer 2 to below 50\% at layer 15, which shows the model loses track of this fundamental concept as it processes deeper.






\section{Discussion and Conclusion}

Our findings reveal a fundamental tension: representations enabling superior performance diverge from human conceptual thinking. Early layers maintain interpretable representations (80-85\% accuracy) that "see" chess as humans do, but deeper layers optimize purely for move prediction (50-65\% accuracy). This isn't a bug but a feature of end-to-end optimization.

Chess960 results are particularly illuminating. The 10-20\% accuracy drop reveals sophisticated pattern matching rather than conceptual understanding. Namely, the model mostly learns specific configurations that correlate with winning, not the conceptual comprehension such as "controlling the center provides mobility." This brittleness suggests models lack compositional understanding for applying principles to novel situations.

\noindent Overall, our contributions include:
\begin{enumerate}
    \item The first Chess960 concept dataset enabling robustness testing;
    \item Evidence that alignment peaks early but deteriorates with depth;
    \item A demonstration that apparent understanding is fragile to creative disruption.
\end{enumerate}

For creative AI, this presents both warning and opportunity. The warning is systems appearing conceptually aligned may operate on different principles. The opportunity: understanding where alignment breaks suggests paths forward, regularization preserving alignment or hybrid architectures separating strategic from tactical reasoning.

\textbf{Acknowledgments:}
This work was supported in part by the US National Science
Foundation AI Institute for Dynamical Systems (dynamicsAI.org)
(grant no. 2112085).

\clearpage


\bibliography{reference}
\bibliographystyle{unsrtnat}
\setcitestyle{numbers} 

\clearpage

\appendix

\section{Technical Appendices and Supplementary Material}
\label{app:math}
\subsection*{A.1 Concept Vector Minimization}

Our first strategy adapts the method from \cite{schut2025bridging}, aiming to find a vector $v \in \mathbb{R}^D$ with minimal L1 norm such that $v^\top r^+ \geq v^\top r^-$ for as many positive–negative board pairs as possible. Here, $r^* \in \mathbb{R}^D$ denotes the activation at layer $l$ for the token used to estimate the move probability: $r^* = z^*_{l,T-1}$. We sample $B$ activation pairs $z^+$ and $z^-$ in a batch, and optimize $v$ via:

\begin{equation}
\mathcal{L}(v) = \lambda \|v\|_1 + \frac{1}{B} \sum_{i=1}^{B} \text{ReLU}(v^\top r^-_i - v^\top r^+_i)
\end{equation}

This promotes separation by maximizing the margin $v^\top r^+_i - v^\top r^-_i$ for each pair while encouraging sparsity in $v$ through the L1 penalty.

\subsection*{A.2 Logistic Regression}

Our second strategy frames the problem as a binary classification task, where the boards containing the concept ($z^+$) are labeled 1 and the boards without the concept ($z^-$) are labeled 0. Similar to the first strategy, we sample a batch of $B$ activation pairs ($z^+, z^-$) and optimize $v \in \mathbb{R}^D$ and $b \in \mathbb{R}$ using $r^*$ via:

\begin{equation}
\mathcal{L}(v, b) = \lambda \|(v, b)\|_1 - \frac{1}{2B} \sum_{i=1}^{B} \left[\log \sigma(v^\top r^+_i + b) + \log(1 - \sigma(v^\top r^-_i + b))\right]
\end{equation}

where $\sigma$ is the sigmoid function. This loss encourages both sparsity in $(v, b)$ and accurate discrimination between $z^+$ and $z^-$. The subsequent predictions are assigned label 1 if $\sigma(v^\top r + b) > 0.5$, and 0 otherwise.

\subsection*{A.3 All Sequence Neural Network}

Our third strategy extends Logistic Regression by using the activations of all tokens in the sequence, $R^* = z^*_l \in \mathbb{R}^{T \times D}$, rather than only the move-token activation at layer $l$. We train a small neural network $S$ to predict the presence of the concept from these activations. We use the objective:

\begin{equation}
\mathcal{L}(v_1, b_1, v_2, b_2) = \lambda \|(v_1, b_1, v_2, b_2)\|_1 - \frac{1}{2B} \sum_{i=1}^{B} \left[\log S(R^+_i) + \log(1 - S(R^-_i))\right]
\end{equation}

where $S: \mathbb{R}^{T \times D} \rightarrow [0, 1]$ is defined as:

\begin{equation}
S(R) = \sigma(v^\top_2 \text{ReLU}(R^\top v_1 + b_1) + b_2)
\end{equation}

with parameters $v_1 \in \mathbb{R}^T$, $b_1 \in \mathbb{R}^D$, $v_2 \in \mathbb{R}^D$, and $b_2 \in \mathbb{R}$. The network takes the activation sequence $s \in \mathbb{R}^{T \times D}$ and outputs a scalar probability. As before, we encourage sparsity in $S$ through the L1 penalty while training it to discriminate between $z^+$ and $z^-$. Predictions are assigned label 1 if $S(s) > 0.5$, and 0 otherwise.

\section{Full Results}
\label{app:res}
\begin{figure}[ht]
  \centering
  \includegraphics[width=\linewidth]{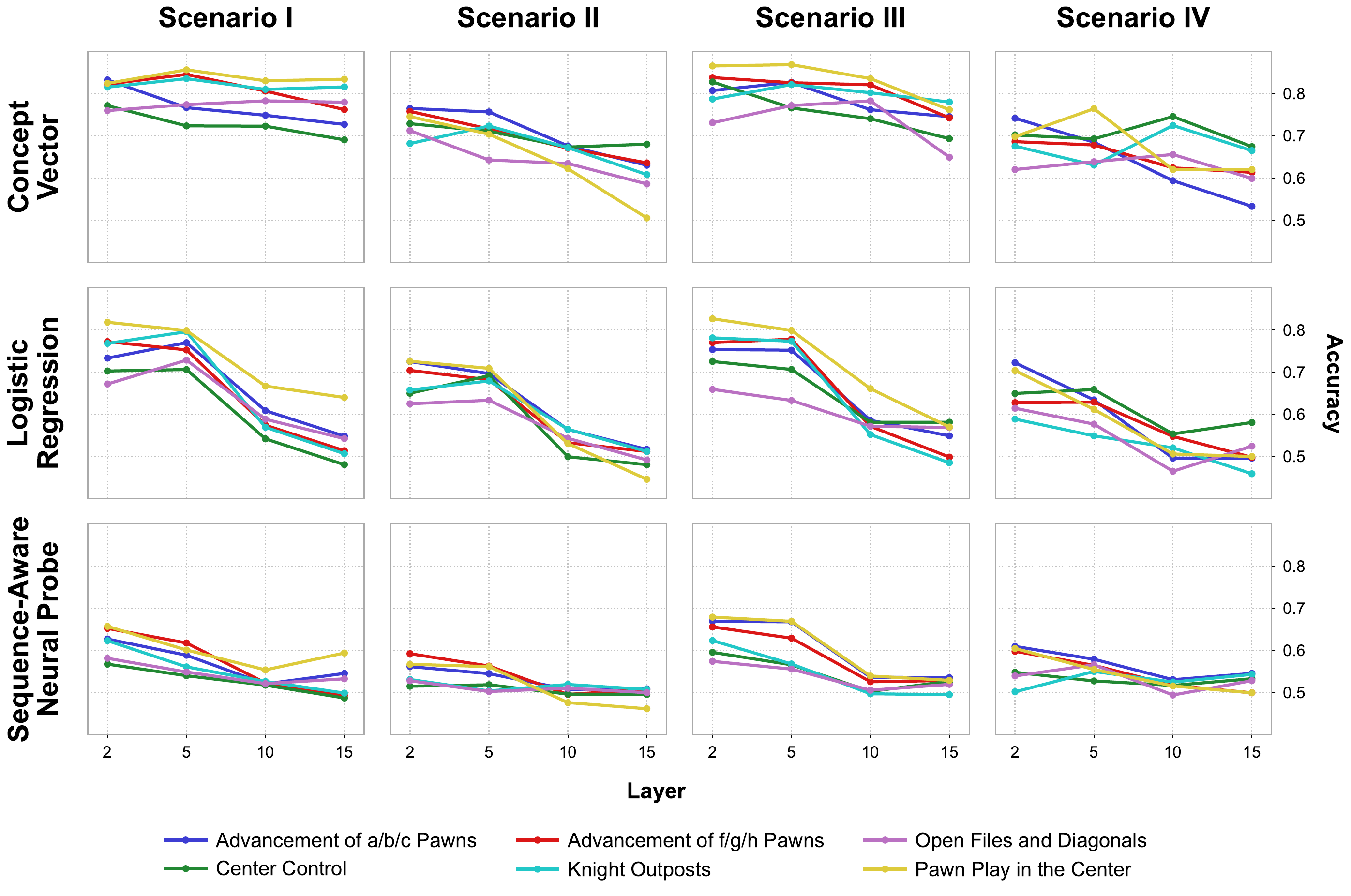}
  \caption{Full results of Figure~\ref{fig:concept_trends}, the overall trend supports the takeaway that the human concepts fade as the network goes deeper.}
  \label{fig:concept_trends_full}
\end{figure}

Here we share the full results from our analysis in Section~\ref{sec:layer-wise-divergence}. Results in Figure~\ref{fig:concept_trends_full} show that the findings generalize across all three probing methods applied. 

\section{Example Concepts}
\label{app:conc}
Here, we provide some example chess concepts from the STS dataset, to give  readers the flavor of the concrete nature of these concepts. 
\begin{figure}[ht]
  \centering
  \includegraphics[width=\linewidth]{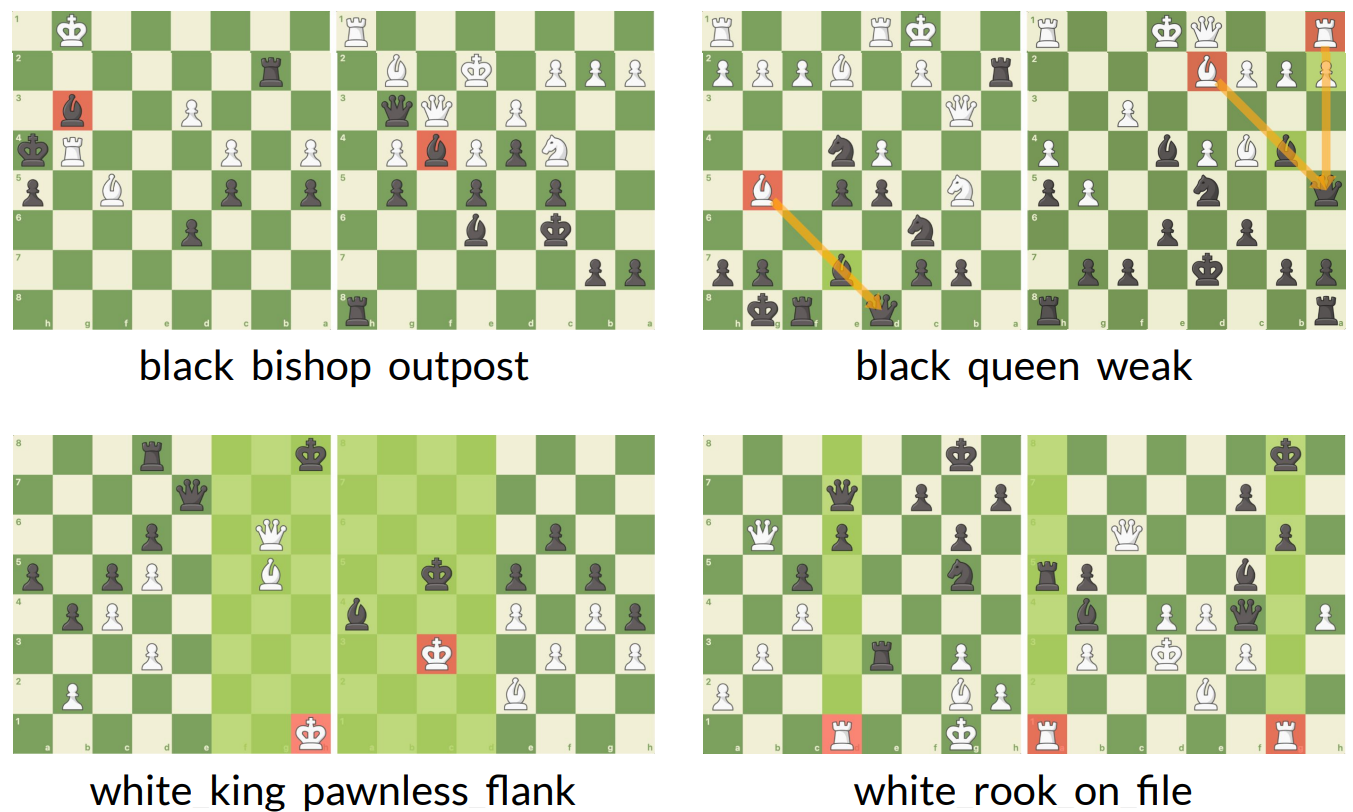}
  \caption{Examples of Human-Concept Categories. The top left illustrates a black bishop outpost, where a black bishop is anchored by pawns deep into the white position. The top right illustrates a black weak queen, where the queen is either directly under attack, or is pinned to a piece. On the bottom left, we have a white king located on a side of the board where neither player has any pawns, and on the bottom right, we illustrate a white rook on a semi-open file, where there is no white pawn opposing it.}
  \label{fig:example_concepts}
\end{figure}

\end{document}

%% file: best_layer_results.tex
\begin{table}[t]
\centering
\footnotesize
\setlength{\tabcolsep}{6pt}
\caption{\textbf{Chess concepts are fragile: accuracy plummets when memorized patterns are disrupted.} Human concept recognition accuracy across three probing methods and six strategic concepts in four scenarios: (I) train/test on standard chess, (II) train standard/test Chess960, (III) combined training, (IV) Chess960 only. The 10-20\% drop from Scenario I to II reveals the model relies on memorization rather than genuine conceptual understanding. Bold values indicate best performance per row.}

\begin{tabular}{l l S[table-format=2.2] S[table-format=2.2] S[table-format=2.2] S[table-format=2.2]}
\toprule
\multicolumn{1}{c}{\multirow{2.5}{*}{\textbf{Concept}}} &
\multicolumn{1}{c}{\multirow{2.5}{*}{\textbf{Method}}}  &
\multicolumn{4}{c}{\textbf{Scenario}} \\
\cmidrule(lr){3-6}
& & \textbf{I} & \textbf{II} & \textbf{III} & \textbf{IV} \\
\midrule
\multirow{3}{*}{Advancement of a/b/c Pawns}
 & Concept Vector & \textbf{83.29} & 76.53 & 82.71 & 74.20 \\
 & Log. Regression     & \textbf{76.99} & 72.50 & 75.37 & 72.22 \\
 & Seq.-Aware N. Probe   & 63.44 & 56.18 & \textbf{67.86} & 62.76 \\
\midrule
\multirow{3}{*}{Advancement of f/g/h Pawns}
 & Concept Vector & \textbf{84.60} & 75.83 & 83.85 & 68.65 \\
 & Log. Regression     & 77.24 & 70.41 & \textbf{77.81} & 62.90 \\
 & Seq.-Aware N. Probe   & 65.34 & 59.24 & \textbf{65.61} & 61.38 \\
\midrule
\multirow{3}{*}{Center Control}
 & Concept Vector & 77.19 & 72.91 & \textbf{82.81} & 74.60 \\
 & Log. Regression     & 70.62 & 69.17 & \textbf{72.52} & 65.87 \\
 & Seq.-Aware N. Probe   & 57.19 & 51.87 & \textbf{59.56} & 56.48 \\
\midrule
\multirow{3}{*}{Knight Outposts}
 & Concept Vector & \textbf{83.60} & 72.36 & 82.18 & 72.49 \\
 & Log. Regression     & \textbf{79.60} & 67.92 & 78.14 & 58.86 \\
 & Seq.-Aware N. Probe   & 62.33 & 53.12 & \textbf{62.35} & 56.02 \\
\midrule
\multirow{3}{*}{Open Files and Diagonals}
 & Concept Vector & \textbf{78.33} & 71.25 & 78.32 & 65.61 \\
 & Log. Regression     & \textbf{72.86} & 63.33 & 65.91 & 61.44 \\
 & Seq.-Aware N. Probe   & \textbf{59.80} & 53.40 & 59.40 & 58.00 \\
\midrule
\multirow{3}{*}{Pawn Play in the Center}
 & Concept Vector & 85.68 & 74.59 & \textbf{86.92} & 76.46 \\
 & Log. Regression     & 81.83 & 72.57 & \textbf{82.66} & 70.37 \\
 & Seq.-Aware N. Probe   & 66.77 & 56.74 & \textbf{69.77} & 62.57 \\
\bottomrule
\end{tabular}
\label{tab:scenario-pivot}
\end{table}